\pdfoutput=1

\documentclass[11pt]{article}

\usepackage[]{emnlp2021}

\usepackage{times}
\usepackage{latexsym}
\usepackage{graphicx}
\usepackage{amsmath}
\newcommand{\mc}{\multicolumn}
\usepackage{booktabs}
\usepackage[export]{adjustbox}
\usepackage{soul}
\usepackage{color}
\definecolor{myyellow}{rgb}{1, 1, 0.6}
\DeclareRobustCommand{\hlyellow}[1]{{\sethlcolor{myyellow}\hl{#1}}}
\definecolor{myred}{rgb}{1, 0.6, 0.6}
\DeclareRobustCommand{\hlred}[1]{{\sethlcolor{myred}\hl{#1}}}
\definecolor{myblue}{rgb}{0.6, 0.7, 1}
\DeclareRobustCommand{\hlblue}[1]{{\sethlcolor{myblue}\hl{#1}}}
\definecolor{mygreen}{rgb}{0.5, 1, 0.5}

\definecolor{myorange}{rgb}{1, 0.8, 0.5}
\DeclareRobustCommand{\hlorange}[1]{{\sethlcolor{myorange}\hl{#1}}}

\usepackage[T1]{fontenc}

\usepackage[utf8]{inputenc}

\usepackage{microtype}

\title{Discovering Explanatory Sentences in Legal Case Decisions Using Pre-trained Language Models}

\author{Jaromir Savelka \\
  School of Computer Science \\
  Carnegie Mellon University \\
  \texttt{jsavelka@cs.cmu.edu} \\\And
  Kevin D. Ashley \\
  School of Law \\
  University of Pittsburgh \\
  \texttt{ashley@pitt.edu} \\}

\begin{document}
\maketitle
\begin{abstract}
Legal texts routinely use concepts that are difficult to understand. Lawyers elaborate on the meaning of such concepts by, among other things, carefully investigating how have they been used in past. Finding text snippets that mention a particular concept in a useful way is tedious, time-consuming, and, hence, expensive. We assembled a data set of 26,959 sentences, coming from legal case decisions, and labeled them in terms of their usefulness for explaining selected legal concepts. Using the dataset we study the effectiveness of transformer-based models pre-trained on large language corpora to detect which of the sentences are useful. In light of models' predictions, we analyze various linguistic properties of the explanatory sentences as well as their relationship to the legal concept that needs to be explained. We show that the transformer-based models are capable of learning surprisingly sophisticated features and outperform the prior approaches to the task.
\end{abstract}

\section{Introduction}
Written laws enacted by legislative bodies set forth the collection of legally binding rules of conduct (e.g., rights, prohibitions, duties). Understanding written laws is difficult because the abstract rules must account for a variety of situations, even those not yet encountered. Written laws communicate general standards and refer to classes of persons, acts, things, and circumstances \cite[p. 124]{hart1994concept}. Therefore, legislators use vague \cite{endicott2000vagueness}, open textured \cite{hart1994concept} terms, abstract standards \cite{endicott2014law}, principles, and values \cite{daci2010legal} to deal with the inherent uncertainty.

For example, let us focus on the two emphasized concepts from the following written provision of law (29 U.S. Code \S\ 203):

\begin{quote}
\footnotesize
    ``Enterprise'' means the \emph{related activities} performed [\textellipsis] for a \emph{common business purpose} [\dots].
\end{quote}

\noindent Understanding of the provision depends on understanding the meaning of the two emphasized concepts. Any doubts about the meaning may be removed by explanation or interpretation \cite{maccormick1991interpreting}. Even small differences in understanding of a single concept may be crucial for determining how a provision applies and what are its effects in a particular context.

For example, the  meaning of the concept \emph{common business purpose} could be crucial in determining if two restaurants in different parts of the same city, sharing a single owner, constitute an ``enterprise.'' The explanation of the concept would involve an investigation of how has it been referred to, explained, interpreted, or applied in the past. This is an important step that enables a lawyer to come up with arguments in support of or against particular accounts of meaning \cite{savelka2015open,savelka2021role}.

Searching through a database of legal documents a lawyer may retrieve sentences such as the following:

\begin{enumerate}
\footnotesize
\item Courts have held that a joint profit motive is insufficient to support a finding of \emph{common business purpose}.
\item The fact of common ownership of the two businesses clearly is not sufficient to establish a \emph{common business purpose}.
\item The third test is ``\emph{common business purpose}.''
\end{enumerate}

\noindent Some of these sentences are most likely useful for explaining the concept (1 and 2) but others appear to have very little value (3). Manually reviewing such sentences is labor intensive.

We would like to rank more highly the sentences the goal or effect of which is to elaborate upon the meaning of the selected concept. These include, but are not limited to, (i) definitional sentences (e.g., a sentence that provides a test for when the concept applies), (ii) sentences that state explicitly in a different way what the concept means or state what it does not mean, (iii) sentences that provide an example, instance, or counterexample of the concept, and (iv) sentences that show how a court determines whether something is such an example, instance, or counterexample. 




\section{Related and Prior Work}
In prior work, we employed a variety of traditional information retrieval (IR) measures and their combinations, e.g., BM25, novelty, topic modeling \cite{savelka2019improving,savelka2020discovering,savelka2021legal}. These turned out to be remarkably successful in finding documents or their parts (e.g., paragraphs) that are likely to contain useful sentences. However, they fell short in performing finer-grained evaluation of the sentences contained in those document parts. Using a learning-to-rank approaches on hand-crafted features led to only moderate improvements \cite{savelka2016extracting,savelka2020learning}. In this work, we show that transformer-based pre-trained models (BERT family) are capable of such fine-grained evaluation by learning to detect sophisticated semantic features in sentences themselves as well as in their relationships to the explained concepts. Furthermore, we show that many of these features are sensible to humans.

The models based on the BERT architecture have been successfully used in a variety of IR tasks. A comprehensive survey of text ranking with transformers, such as BERT, is provided in \cite{lin2020pretrained}. Several simple applications of BERT to ad hoc document retrieval are presented in \cite{yang2019simple}. 
Successful applications of BERT for retrieval of short texts such as sentences are presented in \citet{yilmaz2019cross} and \citet{rao2019bridging}. 
Similar to the utilization of provisions of written law in this work, the authors of \citet{mehrotrampii2019mpii} demonstrated the effectiveness of using a query context in a re-ranking component based on BERT. In \citet{nogueira2019multi} BERT is fine-tuned on query-retrieved document pairs as is done in this work. 

There are examples of successful applications of BERT on legal texts as well. 
A task of retrieving related case-law similar to a case decision a user provides is tackled in~\citet{rossi2019legal}. 
BERT was also proposed as one of the approaches to predict court decision outcomes given the facts of a case \cite{chalkidis2019neural}. BERT has been successfully used for classification of legal areas of Supreme Court judgments \cite{howe2019legal}. BERT was used to tackle the challenging task of case law entailment \cite{rabelo2019combining,westermann2020paragraph}. 
BERT was also used in learning-to-rank settings, as is done in this work, for retrieval of legal news~\cite{sanchez2020easing}. Systematic investigation of BERT's adaptation to the legal domain, resulting in a release of several legal-BERT models, was performed in \cite{chalkidis2020legal}. RoBERTa (variation of BERT) model was used for classification of legal principles applied in court case decisions \cite{gretok2020transformers}. The ability of pre-trained language models (RoBERTa) to generalize beyond the legal domain and dataset they were trained on was analyzed in~\cite{savelka2020cross}.

\section{Data Set}
\label{sec:dataset}
We downloaded the complete bulk data from the Caselaw access project\footnote{A small portion of the dataset is available at \url{case.law}. The complete dataset could be obtained upon entering into research agreement with LexisNexis.} which includes all official, book-published U. S. cases from all federal and state courts as well as from a number of territorial courts \cite{caselaw}. The dataset comprises more than 6.7 million unique cases. 
For document indexing we used a lemmatizer based on the so-called induced ripple-down rules \cite{jurvsic2010lemmagen}.\footnote{\url{http://lemmatise.ijs.si}} Using the U.S. case law sentence segmenter~\cite{savelka2017sentence} we divided each case into individual sentences (0.8 billion).

We queried the system for sentences mentioning 42 selected legal concepts (i.e., terms/phrases, such as ``audiovisual work,'' or ``electronic signature'') coming from provisions of the U.S. Code (the official collection of federal statutes).\footnote{\url{https://www.law.cornell.edu/uscode/text/}} Given the constraints imposed by available resources, we made the best effort to create a well-balanced dataset covering 20 different areas of legal regulation (26,959 retrieved sentences in total).

Eleven law students classified the sentences in terms of four categories with respect to their utility for explaining the legal concepts:

\begin{enumerate}
    \item \textbf{High value} -- This category is reserved for sentences the goal of which is to elaborate on the meaning of the concept.
    \item \textbf{Certain value} -- Sentences that provide grounds to draw some conclusions about the meaning of the concept.
    \item \textbf{Potential value} -- Sentences that provide additional information over what is known from the provision of law.
    \item \textbf{No value} -- Sentences that do not provide any additional information over what is known from the provision.
\end{enumerate}

\noindent Annotators needed to be properly trained to deal with this challenging task. We adopted multiple measures to ensure the annotations of the resulting dataset are of high-quality. The most important one was a second-pass annotation performed by two annotators with a completed law degree ($\alpha=0.79$).\footnote{To measure inter-annotator agreement we used Krippendorff's $\alpha$ \cite{krippendorff2011computing}.}

\begin{figure}[]
    \centering
    \includegraphics[width=0.47\textwidth]{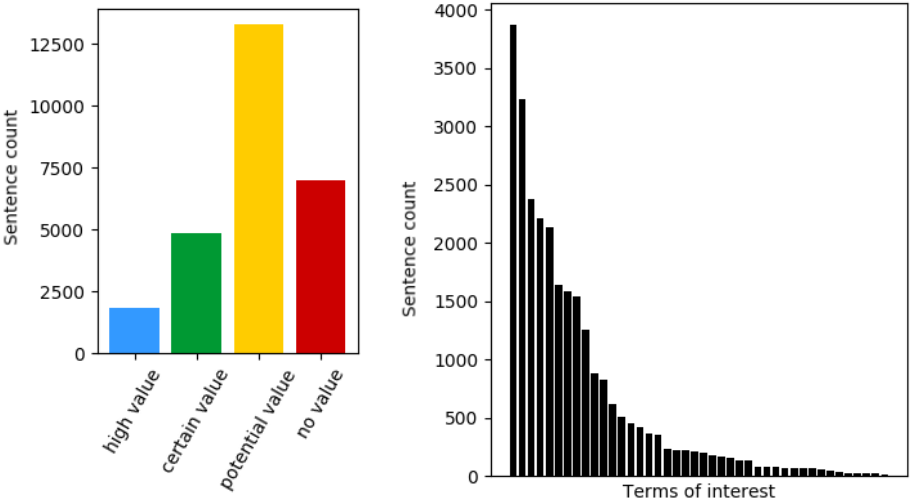}
    \caption{The graph on the left shows the distribution of the labels across all the sentences retrieved for the 42 selected concepts. The graph on the right presents the distribution of the number of sentences retrieved for each concept.}
    \label{fig:data_set_overview}
\end{figure}

Figure \ref{fig:data_set_overview} shows the overall distribution of the labels and the number of sentences associated with each concept/query. The Figure shows that the less valuable categories, `no value' and `potential value,' are dominant. For all the ``larger'' queries and almost all the ``small'' queries it holds that either the `no value' or the `potential value' category is the most numerous one. No matter the size, it is still the case that some of the terms contain quite a considerable number of more valuable sentences (e.g., ``audiovisual work'' or ``switchblade knife'') while others are significantly more limited in this respect (e.g., ``essential step'' or ``hazardous liquid''). As the dataset has not yet been released to the public we are making it available with this paper.\footnote{\url{https://github.com/jsavelka/statutory_interpretation}}

\section{Experiments}
\label{sec:experiments}
In this work, we use RoBERTa---a robustly optimized BERT pretraining approach \cite{liu2019roberta}---as the starting point for the rankers.\footnote{\url{https://github.com/pytorch/fairseq/tree/master/examples/roberta}} Out of the available language models we chose to work with the smaller roberta.base model that has 125 million parameters. This choice was motivated by the ability to iterate the experiments faster when compared to working with roberta.large with 355 million parameters.

We experiment with three different setups. In the first setup we fine-tuned the base RoBERTa model on the task of classifying retrieved sentences in terms of their value for explaining the legal concepts. In prediction we then applied the model to classify the sentences, that were not seen during fine-tuning, in terms of the four value categories (see Section \ref{sec:dataset}). By applying softmax to the final prediction layer we obtained the probability distribution over the four possible classes.
To obtain the sentence's score we compute an inner product between the class probability distribution and value weight vector $(0, 1, 2, 3)^T$. The motivation to use the approach over considering only the predicted class is to take into account the confidence of the prediction. 
Henceforth, this model is referred to as \emph{BERT snt} because it infers the usefulness of a sentence for explaining a legal concept from the sentence only.

In the second setup we fine-tune the base RoBERTa model on the sentence pair classification task. The model is provided a legal concept in place of the first sentence and a retrieved sentence as the second one. The task of predicting sentence value is thus recast as predicting the relationship between the concept and the retrieved sentence. The goal is still to predict one of the four sentence value labels. As in the previous setup, we applied softmax to the final prediction layer to obtain a probability distribution over the classes. Sentences' scores are determined in the same way as well. Henceforth, this model is referred to as \emph{BERT qry2snt}.

The third setup is similar to the second one. Here, we again fine-tune the base RoBERTa model on the sentence pair classification task. Unlike in the second setup, the model is fine-tuned on the whole provision of written law as the first sentence and the retrieved sentence as the second one. Therefore, in this experiment the task is understood as prediction of the relationship between the provision of law and the retrieved sentence. As in the two previous experiments, softmax is applied to the final prediction layer and the probability distribution over the classes is obtained. Henceforth, this model is referred to as \emph{BERT sp2snt}.

In all the experiments, we fine-tuned the base RoBERTa model, with a linear classification layer on top of the pooled output, for 10 epochs on the training splits of the selected datasets. We used the batch size of 8 which is the maximum allowed by our hardware setup (1080Ti with 11GB) given we set the length of a sequence to 512 (maximum). As optimizer we use the Adam algorithm \cite{kingma2014adam} with initial learning rate set to $4e^{-5}$. We stored models' checkpoints after the end of each training epoch. The checkpoints are evaluated on the validation set (see Section \ref{sec:evaluation} for details). The model with the highest F$_1$ on the validation set was then selected as the one to make predictions on the test sets.

\subsection{Evaluation}
\label{sec:evaluation}
Since the notion of relevance in this work is non-binary, we use normalized discounted cumulative gain ($NDCG$) to evaluate the performance of different approaches. An output of the presented ranking algorithms for each concept/query $q_j$ has the form of an ordered tuple of sentences $S_j=(s_1, s_2, \dots, s_n)$. We chose to evaluate the rankings at $k=10$ and 100 which means that the tuples produced by the algorithms are truncated to the respective lengths. Note that the chosen values of $k$ are higher than typical. Measuring at $k=100$ may even appear somewhat extreme. However, legal search differs from the general web search. Assuming a lawyer has confidence in the query (based on seeing several relevant hits towards the top of the results' list), he or she might be inclined to inspect the results way beyond what would a typical web search user do. For each query $q_j$ the $NDCG$ at each $k$ is then computed as:

\begin{equation*}
    NDCG(S_j,k)=\frac{1}{Z_{jk}}\sum_{i=1}^k\frac{rel(s_i)}{log_2(i+1)}
\end{equation*}

\noindent The function $rel(s_i)$ takes a sentence as an input and outputs its value in a numerical form. It is defined as follows:

\begin{equation*}
    rel(s_i)=
    \begin{cases}
    3 & \text{if $s_i$ has high value}\\
    2 & \text{if $s_i$ has certain value} \\
    1 & \text{if $s_i$ has potential value} \\
    0 & \text{if $s_i$ has no value} \\
    \end{cases}
\end{equation*}

\noindent $Z_{jk}$ is a normalizing quantity which is equal to $NDCG(S_j,k)$ where $S_j$ is the ideal ranking. In our case this would mean that all the $s_i$ with `high value' labels are at the beginning positions of the tuple, followed by those with the `certain value,' then `potential value,' and finally `no value' sentences.

We used stratified sampling to distribute the queries into six folds. There are many dimensions along which the result lists associated with the individual queries could be assessed. Two very important ones are the size of the list (i.e., the number of retrieved sentences) and its richness. Richness is a term often used in technology assisted review in eDiscovery. It refers to the prevalence of responsive documents in a collection (result list in case of this work). We adapted the idea for this work by defining a measure that describes the prevalence of valuable sentences in the dataset. First, we assigned a value to a sentence $s_i$ depending on its label on a scale from 0 to 10:

\begin{equation*}
    val(s_i)=
    \begin{cases}
    10 & \text{if $s_i$ has `high value'}\\
    5 & \text{if $s_i$ has `certain value'}\\
    1 & \text{if $s_i$ has `potential value'}\\
    0 & \text{if $s_i$ has `no value'} \\
    \end{cases}
\end{equation*}

\noindent The reason why we used the scale of 0 to 10 is to overcome the dominance of the less valuable sentences. It is important to emphasize that these scores do not reflect the value ratio among sentences with different labels. In order to determine the richness ($R$) of a concept/query $q_j$ we simply computed an average value of the sentence within a results list associated with the concept/query:

\begin{equation*}
R(q_j)=\frac{1}{n}\sum_{i=1}^n val(s_i)
\end{equation*}

Queries with over 550 retrieved sentences are deemed large whereas the rest is considered small. Figure \ref{fig:data_set_overview} (right) shows that this is where the long tail starts. For richness, we chose 2.0 as a cut-off score. The sentences that fall below are dominated by low value sentences. The sentences that fall above are quite rich in higher value sentences. This resulted in the four groups, i.e., small sparse (12 queries), small dense (18), large sparse (6), and large dense (6). Each of the six splits then contain 2 SmSp, 3 SmDs, 1 LgSp, and 1 LgDs sentences.

All the systems are then evaluated using 6-fold cross-validation. In each iteration four folds are used as a training set, one as a validation set, and one as a test set. We obtained two scores (NDCG@10 and NDCG@100) for each of the 42 concepts/queries. We report the unweighted means (i.e., the size of the result list is not taken into account) of each score for the four groups determined by the stratified sampling as well as the Overall performance. 

For testing statistical significance we employ the strategy suggested by \cite{demsar2006statistical} for testing $k$ methods applied to $N$ datasets. In our experiments, we use the NDCG@100 of the Overall group as the evaluation metric to determine statistical significance. Dem\v{s}ar \shortcite{demsar2006statistical} recommends the Friedman test \citep{friedman1937use}, a non-parametric equivalent of the repeated-measures ANOVA. The null-hypothesis states that all the methods (i.e., the assessed ranker and the baselines) are equivalent. In case the null-hypothesis is rejected, we can draw a conclusion that some methods do differ. In order to learn which of them are different, a post-hoc test needs to be conducted. We use the Holm-Bonferroni step down method \cite{holm1979simple} where the comparisons are performed in  sequential order from the most significant hypotheses until a null-hypothesis that cannot be rejected is encountered. 

\begin{table*}[]
    \centering
    \scriptsize
    \setlength{\tabcolsep}{4.5pt}
    \caption{The table shows the results of the experiments with pre-trained language models. The NDCG@10 and NDCG@100 are shown for the small sparse queries (SmSp), small dense queries (SmDs), large sparse queries (LgSp), large dense queries (LgDs), and all of them together (Overall).}
    \label{tab:lm_results}
    \begin{tabular}{lcccccccccc}
    \toprule
    &\mc{2}{c}{SmSp}&\mc{2}{c}{SmDs}&\mc{2}{c}{LgSp}&\mc{2}{c}{LgDs}&\mc{2}{c}{Overall}\\
    Method              & @10       & @100      & @10       & @100      & @10       & @100      & @10       & @100      & @10       & @100 \\
    \midrule
    Random              &$.38\pm.10$&$.67\pm.15$&$.52\pm.07$&$.76\pm.09$&$.25\pm.16$&$.29\pm.18$&$.47\pm.09$&$.48\pm.09$&$.43\pm.13$&$.63\pm.21$\\
    BM25                &$.47\pm.13$&$.74\pm.11$&$.60\pm.18$&$.79\pm.11$&$.44\pm.21$&$.37\pm.22$&$.61\pm.17$&$.56\pm.12$&$.54\pm.18$&$.68\pm.20$\\
    BM25-c              &$.48\pm.12$&$.76\pm.09$&$.59\pm.17$&$.80\pm.11$&$.49\pm.14$&$.42\pm.17$&$.63\pm.19$&$.55\pm.13$&$.55\pm.16$&$.70\pm.18$\\
    \midrule
    BMp+NW+LDA          &$.55\pm.11$&$.78\pm.12$&$.64\pm.14$&$.82\pm.10$&$.58\pm.16$&$.56\pm.02$&$.65\pm.23$&$.62\pm.11$&$.61\pm.15$&$.74\pm.14$\\
    RF-PWT              &$.60\pm.16$&$.81\pm.11$&$.66\pm.12$&$.83\pm.10$&$.71\pm.17$&$.68\pm.08$&$.67\pm.10$&$.64\pm.09$&$.65\pm.14$&$.77\pm.12$\\
    \midrule
    BERT snt            &$.50\pm.18$&$.71\pm.17$&$.61\pm.14$&$.80\pm.11$&$.46\pm.24$&$.47\pm.21$&$.83\pm.15$&$.77\pm.12$&$.59\pm.20$&$\mathbf{.72\pm.18}$\\
    BERT qry2snt        &$.59\pm.23$&$.76\pm.19$&$.72\pm.18$&$.85\pm.10$&$.64\pm.34$&$.50\pm.28$&$.86\pm.25$&$.77\pm.18$&$.69\pm.24$&$\mathbf{.77\pm.20}$\\
    BERT sp2snt         &$.57\pm.19$&$.80\pm.12$&$.74\pm.15$&$.87\pm.07$&$.73\pm.12$&$.59\pm.18$&$.89\pm.16$&$.80\pm.14$&$.71\pm.19$&$\mathbf{.80\pm.14}$\\
    \bottomrule
    \end{tabular}
\end{table*}

\subsection{Baselines}
\label{sec:baselines}
As baselines we report the performance of a Random system on a large sample of repeated runs (for reference) as well as two simple methods based on BM25. The first method is the Okapi BM25 function \cite{robertson2009probabilistic} applied to query-sentence pairs. The second BM25-based baseline (BM25-c) is a linear interpolation of BM25 applied to the \emph{query-sentence pair} (s) and to the \emph{whole provision of written law-whole case decision pair} (c) it comes from (context). The two BM25 baselines are very close to what is typically used in many legal IR systems. Furthermore, they are very effective baselines that are often not easy to outperform. For comparison, we also report the performance of the best systems from the prior work. \cite{savelka2019improving,savelka2020learning, savelka2021legal}

\section{Results}
The results of the experiments described in Section \ref{sec:experiments} are reported in Table \ref{tab:lm_results} (group and overall means). The top section of the table presents the performance of the three baselines. The two BM25 baselines clearly outperform the Random system. Despite their similar performance they benefit from completely different phenomena. Intuitively, BM25 ranks high sentences that contain multiple mentions of the concept. In this work the method is optimized in such a way that the documents are not penalized for their length. Hence, the system would often prefer very long sentences. Obviously, such a simple approach works to a certain extent. BM25-c is a combination (linear) of the plain BM25 and another BM25 measure applied to the whole text of a case (i.e., sentence's context). Hence, this system can additionally use the fact of the concept appearing many times within the whole text. This is useful because a decision that mentions the term many times is more likely to contain useful sentences than a decision that mentions it just once. Apparently, the BM25-c is the most competitive of the three baselines.

The middle section of Table \ref{tab:lm_results} shows the performance of the two best models from prior work \cite{savelka2019improving, savelka2020discovering}. The BMp+NW+LDA is a linear combination of BM25 applied on a paragraph level, novelty measure, and topic similarity measure. The RF-PWT is the random forest model trained on the 161 hand-crafted features proposed in \citet{savelka2020discovering,savelka2020learning}. These models appear to be an improvement over the two baselines.

The performance of the methods based on the pre-trained language models is very promising. Even the performance of the model that  considers just the sentence itself (\emph{BERT snt}) and completely ignores  the legal concept or the source provision shows promise. The statistical evaluation corroborates the summary statistics reported in Table \ref{tab:lm_results}. The strongest conclusion as to outperforming the two BM25 and the Random baseline can be reached for the \emph{BERT sp2snt} model ($p=0.0002$). While for the \emph{BERT qry2snt} ($p=0.012$) and \emph{BERT snt} ($p=0.022$) models the conclusion is not as strong, it still solidly supports the finding (especially considering the relatively limited size of the dataset). The models also appear to improve over the prior state-of-the-art.

\section{Discussion}
\label{sec:discussion}
While it should be apparent that the sentence detached from the legal concept (and the provision of law it is embedded in) does not provide reliable grounds for determining its value, it appears that the sentences themselves do carry some signal. This is evidenced by the performance of the \emph{BERT snt} model that only considers sentences themselves. This model outperforms the BM25-based baselines. Interestingly, the system correctly recognized that very short pieces of text that do not form full grammatical sentences typically have very little value. For example, the following sentences have been placed at the bottom of their respective rankings (the explained context is highlighted in yellow):

\begin{quote}
\footnotesize
Communication \& \hlyellow{Navigation Equipment}\\
\hlred{[No value]}\vspace{.2cm}\\
B. Non-Disclosure of \hlyellow{PreExisting Works}\\
\hlred{[No value]}
\end{quote}

\noindent Furthermore, it appears that the system relies on features such as the presence of numbering in the sentence, complicated sentence structures, abrupt endings or starts of the sentences, and references, to recognize quotations of written provisions of law and assign a low value to such sentences:

\begin{quote}
\footnotesize
Derives \hlyellow{independent economic value}, actual or potential, from not being generally known to the public or to other persons who can obtain economic value from its disclosure or use; and [f] (2)\\
\hlred{[No value]}
\end{quote}


\noindent This strategy makes sense in general. The quotation could either be the citation of the source provision (`no value') or a citation of a different provision (high chance of different meaning and hence lower value of a sentence). However, there are situations in which the strategy does not work well.

\begin{quote}
\footnotesize
The Electronic Communications Privacy Act of 1986 (ECPA), Pub. L. 99-508, \S 101(a)(6)(C), 100 Stat. 1848, 1849 (1986), codified, as amended, at 18 U.S.C. \S 2510(18) (1986), defines ``\hlyellow{aural transfer}'' to mean ``a transfer containing the human voice at any point between and including the point of origin and the point of reception.''\\
\hlblue{[High value]}
\end{quote}

\noindent The `aural transfer' is a rare example of a concept for which there is a legal definition. As a result \emph{BERT snt} underperforms the Random baseline on this particular concept (NDGC@100 0.53 vs 0.62).

\emph{BERT snt} also seems to have developed a certain tendency to rank high sentences where something is claimed to be something else:


\begin{quote}
\footnotesize
Screen output is considered an \hlyellow{audiovisual work} that falls within the subject matter of copyright.\\
\hlblue{[High value]}
\end{quote}

\noindent This also appears to be a good strategy that works well many times but not always. For example, the following sentence is just `potential value' because it uses ``navigation equipment'' in a different meaning (avionics instead of seafaring):

\begin{quote}
\footnotesize
Avionics are aircraft radios and \hlyellow{navigation equipment}.\\
\hlorange{[Potential value]}
\end{quote}

The above examples demonstrate how the pre-trained deep architecture detects very complex features. It would be quite difficult for a human expert  to  hand-craft such features. While it is not difficult to come up with features such as sentence length, it is far more difficult to come up with features capturing complicated sentence structures, abrupt endings, or subsumption. It is even more difficult to ensure that all the relevant phenomena are considered.

\emph{BERT qry2snt} models the relationship between the legal concept and the retrieved sentences. It appears to perform better than the base \emph{BERT snt} model.  \emph{BERT qry2snt} has access to the same kind of strategies as \emph{BERT snt}, but since it does not ignore the concept it can go further. For example, there is a clear trend of ranking as very high sentences that contain the concept surrounded by quotation marks:

\begin{quote}
\footnotesize
The first subsection of that provision, entitled ``\hlyellow{Navigation Equipment},'' requires tankers to possess global positioning system (``GPS'') receivers, as well as two separate radar systems.\\
\hlblue{[High value]}\vspace{.2cm}\\
We believe the common meaning and general understanding of the term ``\hlyellow{switchblade knife}'' is a knife in which the blade extends and is securely locked open upon the pressing of a button or other mechanism.\\
\hlblue{[High value]}
\end{quote}


\noindent This appears to be a viable strategy. However, there could be instances where it does not work perfectly.

\emph{BERT qry2snt} appears to have the ability to recognize certain linguistic relationships between the term of interest and other parts of a sentence. The following sentences were not recognized as valuable by \emph{BERT snt} but they are correctly ranked very high by \emph{BERT qry2snt}:

\begin{quote}
\footnotesize
Airplanes need wings to fly, but that does not mean that all wing designs have \hlyellow{independent economic value}.\\
\hlblue{[High value]}\vspace{.2cm}\\
As explained above, the duty titles in this case do not qualify as \hlyellow{identifying particulars}.\\
\hlblue{[High value]}\vspace{.2cm}\\
And ``motion pictures'' are ``\hlyellow{audiovisual works} consisting of a series of related images which, when shown in succession, impart an impression of motion, together with accompanying sounds, if any.''\\
\hlblue{[High value]}
\end{quote}


\noindent All these examples seem to exhibit certain higher level patterns that are intuitively very appealing. Rewriting the above sentences into such patterns could look like this:

\begin{quote}
\footnotesize
    {[\ldots]} NOUN\_PHRASE have CONCEPT\vspace{.2cm}\\
    {[\ldots]} qualify [\ldots] NOUN\_PHRASE [\ldots] CONCEPT\vspace{.2cm}\\
    NOUN\_PHRASE is defined to be CONCEPT [\ldots]\vspace{.2cm}\\
    {[\ldots]} ``NOUN\_PHRASE'' are ``CONCEPT [\ldots]''
\end{quote}

\begin{figure*}[]
    \centering
    \includegraphics[width=0.32\textwidth,valign=t]{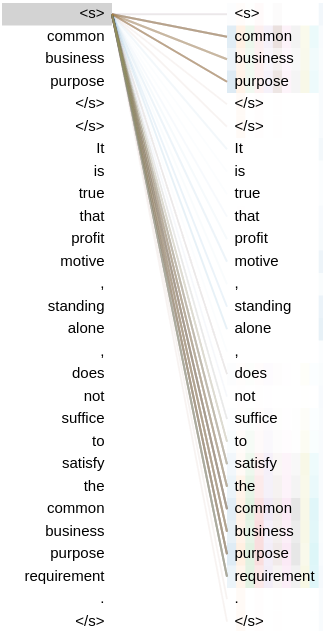}\hspace{.1cm}
    \includegraphics[width=0.32\textwidth,valign=t]{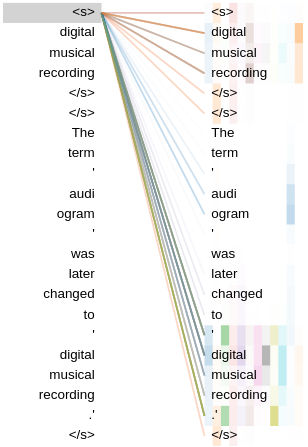}\hspace{.1cm}
    \includegraphics[width=0.32\textwidth,valign=t]{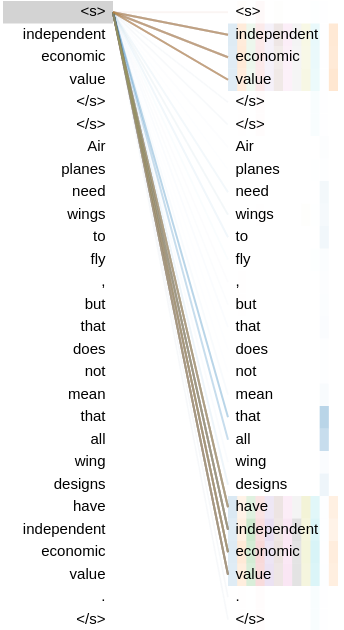}
    \caption[Topic Similarity Threshold Analysis]{The figure provides some indication of what input elements influence the representation that is being used in the final classification step. The model attends to parts of the sentences that really appear to be suggestive about the higher value of a sentence (i.e., ``to satisfy the common business purpose requirement'', the quotation marks surrounding the digital musical recording, or ``that all \ldots have independent economic value'').} 
    \label{fig:bert_weights}
\end{figure*}

\noindent This is corroborated by the inspection of the model weights as applied to several sentences shown in Figure \ref{fig:bert_weights}. The visualization was created using the tool published with \cite{vig2019multiscale}. As mentioned earlier BERT is based on the transformer model from \cite{vaswani2017attention}. An  advantage of using the attention-based model is that it can be interpreted via inspection of the weights assigned to different input elements. As Vig \shortcite{vig2019multiscale} warns one needs to be very conservative with respect to drawing any conclusions. The three diagrams in Figure~\ref{fig:bert_weights} show how much attention  the first special tokens pay to the individual words in the three input sequences. Note that the input sequences each consist of a term of interest and a retrieved sentence. The reason why the first special token is interesting is that this token stands for the vector representing the sequence which is then fed into a classifier. Hence, the visualization provides some indication of what influences the representation that is being used in the final classification step.

All three examples show that \emph{BERT qry2snt} establishes the relationship between the term of interest (first part of the sequence) and its mention in the sentence. Additionally, the model attends to parts of the sentences that appear to be suggestive about the higher value of a sentence (i.e., ``to satisfy the common business purpose requirement'', the quotation marks surrounding the digital musical recording, or ``that all \ldots have independent economic value'').

Finally, the \emph{BERT sp2snt} model that focuses on the relationship between a written provision of law and a retrieved sentence appears to perform better than the \emph{BERT qry2snt model}. This may seem somewhat surprising because \emph{BERT sp2snt} does not have access to the focused legal concept. On the other hand, it is provided with the full provision of law. While \emph{BERT sp2snt} appears to lack the ability of \emph{BERT qry2snt} to detect the useful linguistic patterns attached to the legal concepts, it has the ability to recognize the sentences with `no value' with a high level of accuracy. For example, \emph{BERT qry2snt} ranked the following sentences high:

\begin{quote}
\footnotesize
In that article, a ``wire communication'' is defined as ``an \hlyellow{aural transfer} made in whole or in part through the use of facilities for the transmission of communications by the aid of wire, cable, or other like connection between the point of origin and the point of reception.''
\hlred{[No value]}\vspace{.2cm}\\
The \hlyellow{semiconductor chip product} in turn is defined as: the final or intermediate form of any product--\\
\hlred{[No value]}
\end{quote}


\noindent While these sentences appear to offer valuable definitions of the legal concepts, they merely quote the provision of law, and thus have `no value.' Overall, it appears that with respect to the NDCG scores, it is extremely important to make sure that sentences such as these do not appear at the top positions of the rankings.

\begin{figure}[]
    \centering
    \includegraphics[width=0.4\textwidth]{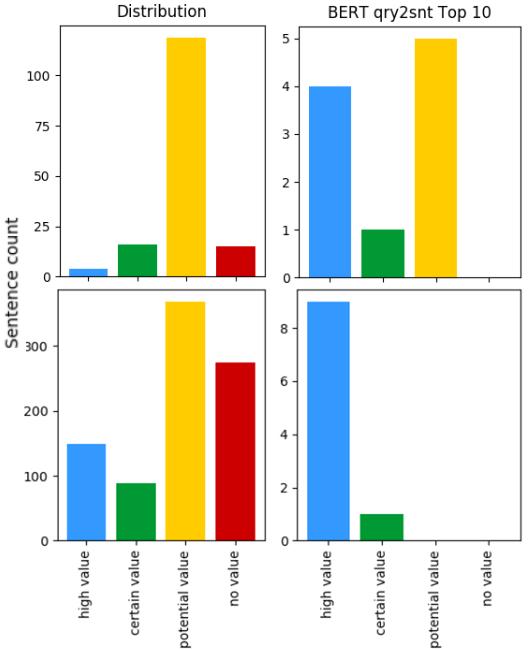}
    \caption{The top two graphs show the sentence value distribution for the concept ``navigation equipment.'' The graph on the left shows the overall distribution while the graph on the right shows the distribution of the top ten sentences retrieved by \emph{BERT qry2snt}. The bottom two graphs show the same for the concept of ``common business purpose.''}
    \label{fig:top10_dist}
\end{figure}

Finally, to provide some concrete examples of the rankings produced by the assessed models Figure \ref{fig:top10_dist} shows the distributions of labels of the top 10 retrieved sentences as compared to the overall distribution for two selected concepts (``navigation equipment'' and ``common business purpose''). The changes in the distributions demonstrate how effective the models can be.

\begin{figure}[]
    \centering
    \includegraphics[width=0.48\textwidth]{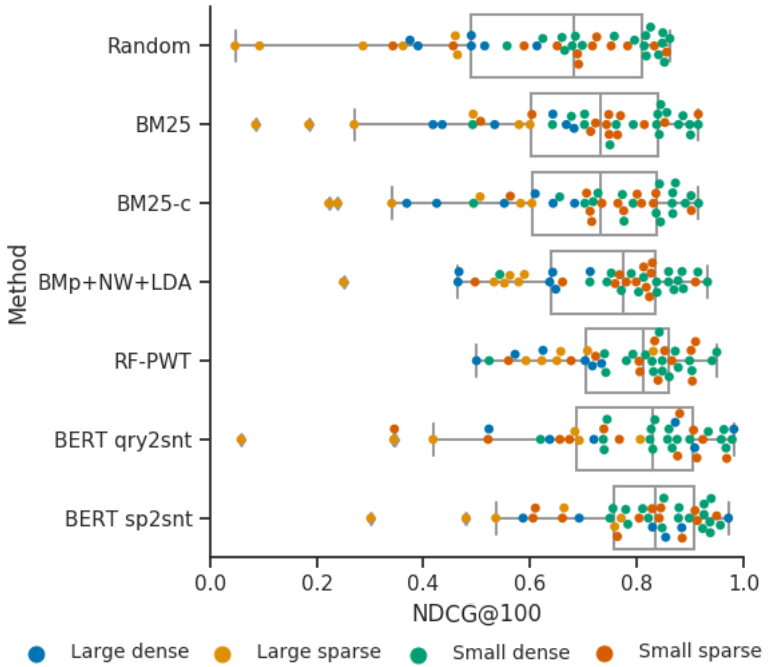}
    \caption{The figure shows scatter plots of the performance on the individual 42 queries measured in terms of NDCG@100 showing the progression from application of simpler similarity methods towards more complex learning-to-rank systems.}
    \label{fig:overview}
\end{figure}

Figure \ref{fig:overview} shows box and whisker plots augmented with swarm plots of per query performance for the evaluated systems. Interestingly, it appears that the progression starting from the BM25 method and ending with the RF-PWT (i.e., the prior work referenced above) mostly improves the performance by correcting the disastrous performance of the queries on the left tail of the swarm plots. Despite certain improvements happening at the right side as well, these are dwarfed by the events on the left.

The pre-trained language models fine-tuned on the task of sentence pair classification are interesting because they no longer focus on the improvement of the lowest performing queries only. They also bring notable improvement to the queries where the performance had already been decent. This is especially true for the BERT qry2snt model that is completely oblivious to the source provision. Hence, this model cannot address the requirement of ``providing additional information'' as well as the requirement of ``using the term of interest in the same meaning.'' Indeed, it appears that the model has similar issues with a number of queries that the BM25 method had. Yet, despite these notable issues the overall performance is comparable to (if not better than) the RF-PWT.

The BERT sp2snt method uses the source provision instead of the term of interest. On closer inspection one sees three large sparse queries that are not handled well by this method in Figure \ref{fig:overview}. There are two reasons why a sentence could have `no value.' It either provides no additional information or it uses the term in a completely different meaning. The three mishandled queries have many sentences that use the term in a different meaning. It appears that BERT sp2snt learned to down-rank the sentences that do not provide additional information quite reliably whereas it completely failed to learn to down-rank the sentences that use the term in a different meaning. The data set may be too small for the method to capture this aspect.

\section{Conclusions and Future Work}
In this work, we showed that pre-trained language models based on transformers can be fine-tuned for the special task of retrieving sentences for explaining legal concepts. Specifically, we demonstrated that a pre-trained RoBERTa base model, fine-tuned on three variations of the task, resulted in effective ranking functions outperforming the BM25 baselines. The promising performance of  \emph{BERT snt} reveals the interesting fact that sentences themselves carry certain signal about their usefulness. The even stronger performance of \emph{BERT qry2snt} and \emph{BERT sp2snt} points to the important interactions among a legal concept, the provision of law in which it is embedded, and retrieved sentences, that both need to be accounted for in order to perform well in this challenging task. The whole work  demonstrates  the effectiveness of methods based on pre-trained language models applied to a legal domain task. This is important because advances in general NLP and ML do not always transfer in a straightforward manner to specialized domains such as automatic processing of legal or medical texts. Importantly, we fill the gap in prior work by showing that the transformer based methods are capable of fine-grained evaluations of the individual sentences as to their usefulness.

The application of pre-trained language models to the task of discovering sentences explaining legal concepts yielded promising results. At the same time, the work is subject to limitations and leaves much room for improvement. Hence, we suggest several directions for future work:

\begin{itemize}
    \item Focus on  \emph{diversity} in addition to  \emph{relevance} to ensure that the top results do not  repeat the same sentences.
    \item Account for all  three constituents, i.e., the legal concept, the written provision of law, and retrieved sentences, simultaneously. 
    \item Investigate the effects of including the context of a retrieved sentence, i.e., the full text of a case decision.
    \item Invest more resources in developing and \emph{extending} the \emph{dataset}.
    \item Investigate retrieving sentences  from other types of legal documents beyond court case decisions (e.g., legislative histories, commentaries).
    \item Perform an \emph{extrinsic evaluation} of the system in the context of an end-to-end legal project.
\end{itemize}

\section*{Acknowledgements}
The first author would like to acknowledge the University of Pittsburgh as his home institution during the time this work was conducted. This work was supported in part by a National Institute of Justice Graduate Student Fellowship (Fellow: Jaromir Savelka) Award \# 2016-R2-CX-0010, ``Recommendation System for Statutory Interpretation in Cybercrime,'' and by a University of Pittsburgh Pitt Cyber Accelerator Grant entitled ``Annotating Machine Learning Data for Interpreting Cyber-Crime Statutes.''


\bibliography{anthology,custom}
\bibliographystyle{acl_natbib}




\end{document}